\documentclass{article}
\usepackage{spconf,amsmath,epsfig}

\usepackage{wrapfig}
\usepackage{algorithm}
\usepackage{algorithmic}
\usepackage{array}

\usepackage{amsmath}
\usepackage{amssymb}
\usepackage{lineno,hyperref}
\usepackage{booktabs}
\usepackage{multirow}
\usepackage{graphics}
\usepackage{graphicx}
\usepackage{subfigure}
\usepackage{float}
\usepackage{color}
\usepackage{diagbox}
\begin{document}\sloppy
%
\def\x{{\mathbf x}}
\def\L{{\cal L}}

\title{Audio2Face: Generating  Speech/Face Animation from Single Audio  with Attention-Based Bidirectional LSTM Networks}
%
\name{Guanzhong Tian$^1$ , Yi Yuan$^2$ , Yong Liu$^1$ }
\address{}
\address{$^1$ Institute of Cyber-systems and Control, Zhejiang University, $^2$ Fuxi AI Lab, Netease}

\maketitle

\begin{abstract}
We propose an end to end deep learning approach for generating real-time facial animation from just audio. Specifically, our deep architecture  employs deep bidirectional long short-term memory network and attention mechanism to discover the latent representations of time-varying contextual information within the speech and recognize the significance of different information contributed to certain face status. Therefore, our model is able to drive different levels of facial movements at inference and automatically  keep up with the corresponding pitch and latent speaking style in the input audio, with no assumption or further human intervention. Evaluation results show that our method could not only generate accurate lip movements from audio, but also successfully regress the speaker's time-varying facial movements.
\end{abstract}

\begin{figure*}[h]
\centering
\includegraphics[width=1.0\textwidth]{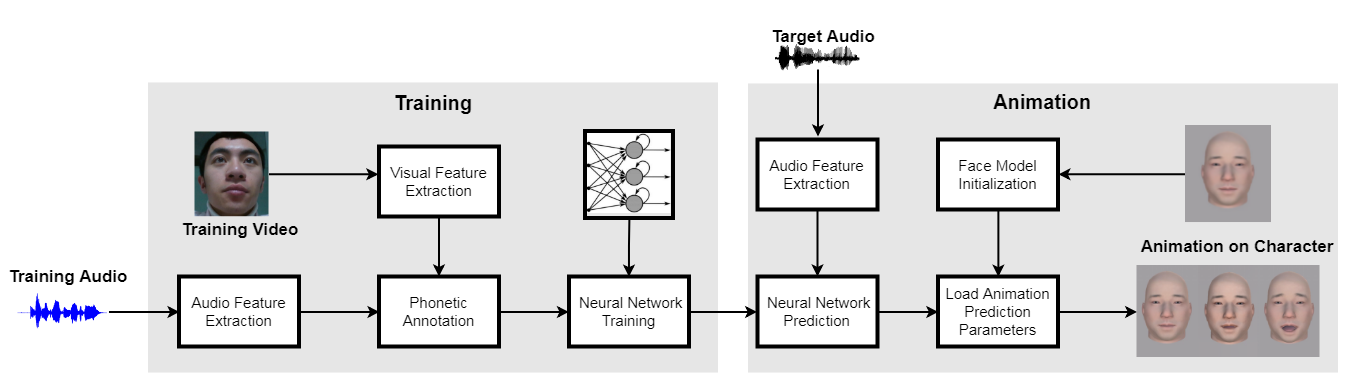}
\caption {Our proposed end-to end deep learning approach to learn a map from phoneme labels to facial/speech animation.}
\label{pipeline}
\end{figure*}
\section{Introduction}
\label{sec:intro}

 As an essential part of human machine interaction, facial animation has been applied on many applications, such as computer games,  animated movies, man-machine conversation, virtual human agents, etc. Currently, great success has been achieved by vision-based facial capture techniques in synthesizing virtual human face. Vision-based performance capture rigs usually take advantage of  motion sensors/markers and drive the animated face with the captured motion of a human face. Although the capture systems are able to provide a pretty acceptable results, active capture rigs tend to be expensive and time-consuming to use because of the elaborate setups and processing steps. Considering the constrains of facial capture methods, other input modalities, such as audio, may be used as a indispensable complementarity to infer facial motions. As a matter of fact, a growing number of animated film now use vision capture systems to produce just key animation, and rely on approaches based on audio or transcript to produce the vast majority of detail.

Although inferring facial motions and expressions from speech is a fundamental human reasoning functionality, it is extremely challenging for computer. On one hand, real-time applications such as inter-person telecommunication requires strong adaptability of wide variability in user voices  and stringent latency requirements. On the other hand, using just audio track to generate facial shapes seems unmanageable due to the inherent uncertainty of the problem— different facial expression may comes from exactly the same audio~\cite{Petrushin1998}. However, we try to address this challenge by using deep neural network, which has show the effectiveness and strong recovery capability by its widely spread on super resolution and image generation problems.

In this work, our primary goal is to recreate a plausible 3D virtual talking avatar that can make reasonable lip movements and then further generate natural facial movements. Our approach is designed exclusively on the basis of audio track, without any other auxiliary input such as images, which makes it increasingly challenging as we attempt to regress visual space from vocal sequence. Another challenge is that facial motions involve multifold activations of correlated regions on the geometric surface of face, which makes it difficult to generate life-like and consistent facial deformations for avatars. To recover as much details from the original input face with low latency, we adopt blendshape models to output the activation parameters of a fixed-topology facial mesh model. First, we extract handcrafted, high-level acoustic features, i.e. mel-frequency cepstral coefficients (MFCCs), from  raw audio as the input to our network. Second,  in order to map the long range context of the  audio signal to effective face shape parameters, we employ bidirectional long short-term memory neural networks with attention mechanism  to model meaningful acoustic feature representations, taking advantage of the locality and shift invariance in the frequency domain of audio sequence. Moreover, we combine attention embedded LSTM layers with fully-connected layers into an end-to-end framework, which aims at generating  temporal transition of lip/facial movements, as well as spontaneous actions.

\section{RELATED WROK}

\textbf{Face/Speech Animation Synthesis} Quite a few methods have been developed to synthesize a face model from either transcript~\cite{Cosatto2003}~\cite{Wang2007} or speech ~\cite{Wang2015}~\cite{Xie2007}. Prior to the advent of deep learning, most of the work was based on analyzing explicit knowledge about the phoneme content of languages, and then build a map between phonemes and target output face model coefficients. The animated movements is then generated by the counterpart coefficients of input phonemes.  There  exists  weaknesses of the linguistic-based method: the lack of a automatic way to generate facial animation from different languages and the incapability for responsing to non-phoneme audio inputs. Due to these inherent constrains and the arise of deep learning models, neural networks have been successfully applied to speech synthesis~\cite{Qian2014}~\cite{Ze2013}
and facial animation~\cite{Ding2015}~\cite{Fan2016}~\cite{Zhang2013} with superior performance in recent years. The reason for the effectiveness of deep neural networks work lies in the capability of extracting the internal correlation and long-term relationship of high-dimensional input data, as well as the highly non-linear mapping from input features to target output space. Fan et al.~\cite{Fan2016} suggest to combine acoustic and text features to acquire AAM (active appearance model) coefficients of the mouth area, which then will be used to drive a life-like talking head. Taylor et al.~\cite{Taylor2017} introduce a  deep neural network(DNN) model to automatically regress AAM coefficients from input phonemes,  which can be retargeted to drive lip movements for different 3D face models.

\textbf{Recurrent Neural Networks}
Recurrent neural network (RNN) ~\cite{Williams1989} is a mutant from traditional neural networks and mainly address the issue of sequence modeling. It is designed with loops to possess the ability of allowing information to persist and integrating temporal contextual information. Due to the problem of vanishing gradients, a new building unit for RNN called long short-term memory(LSTM)~\cite{Graves1997} appears and extend the memory of RNNs. In our work, we design the framework based on assumption that animation frames are connected and a model with longer memory could provide more details for the modeling of facial movements. Therefore, wo choose LSTM network since it is born to help memorize long source input sequence in neural machine mapping.

\textbf{Attention Mechanism}
Attention is, to some extent, motivated by how we pay attention to different regions of an image or correlate words in one sentence ~\cite{Mnih2014}. It has been applied in deep learning, for speech recognition,  image caption and sentence summarization modeling~\cite{Chorowski2015}~\cite{Show2015}~\cite{Rush2015}, etc. A dilemma we are facing when design the network is  that not all extracted features from audio input contribute to the output animation vectors equally. Therefore, the attention mechanism is an ideal and important complement for our architecture, which can be used to train the model to learn to pay selective attention to audio sequence representations.

\section{APPROACH }
In this section, we introduce the architecture of our network, along with details on bidirectional LSTM and audio feature extraction. In order to fulfill our goal of generating  plausible 3D facial/speech animation, we formulate three requirements for our synthesis approach: Lifelike, Real-time and Retargetable. Lifelike (Requirement 1) means the generated animations should be with high fidelity and accurately reflect complex speaking patterns present in visible speech motion. Real-time (Requirement 2) requires the system  to generate animations with low latency. Retargetable (Requirement 3) expects the system to be able to retarget the generated animations to other character rig. To satisfy all of the requirements, we first take a vision-based rigs to capture the complex facial movements of actors and build an appropriate training set. Then we build several 3D blendshape-based cartoon face models with counterpart parameters to control different parts of the face: eyebrows, eyes, lip, jaw, etc.  Further, we bring attention mechanism into bidirectional LSTM networks to map input audio feature to animation parameters. Fig. \ref{pipeline} depicts the pipeline of our system.

\subsection{END-TO-END NETWORK ARCHITECTURE}
 Input a short span of audio, the goal of our framework is to correspondingly return a series of successive lip/facial movements. Our neural network first take in audio feature as input and feed it into  bi-directional LSTM layers for high level feature extraction, then the  attention  layer can learn to pay attention to only the main relevant features and output abstract representations. Finally, we feed these abstract representations to an output network to generate the final 51 dimensional vectors to control the 3D face model. The output network is composed of a pair of dense layers that are used as linear transformer on the data. The first dense layer maps the set of input features to a linear time-varying basis vectors, and the second layer outputs the final blendshape parameter vectors as a weighted sum over the corresponding basis vectors. After the network is trained, we animate the face model by sliding a window over the vocal audio track and evaluate the network at every time step. Due to the bi-LSTM network and the embedded attention mechanism, the network itself has a memory of past input audio features and choose which one can affect the current animation frame.

\subsection{Bidirectional LSTM}
\begin{figure}[h]
\centering
\includegraphics[width=0.5\textwidth]{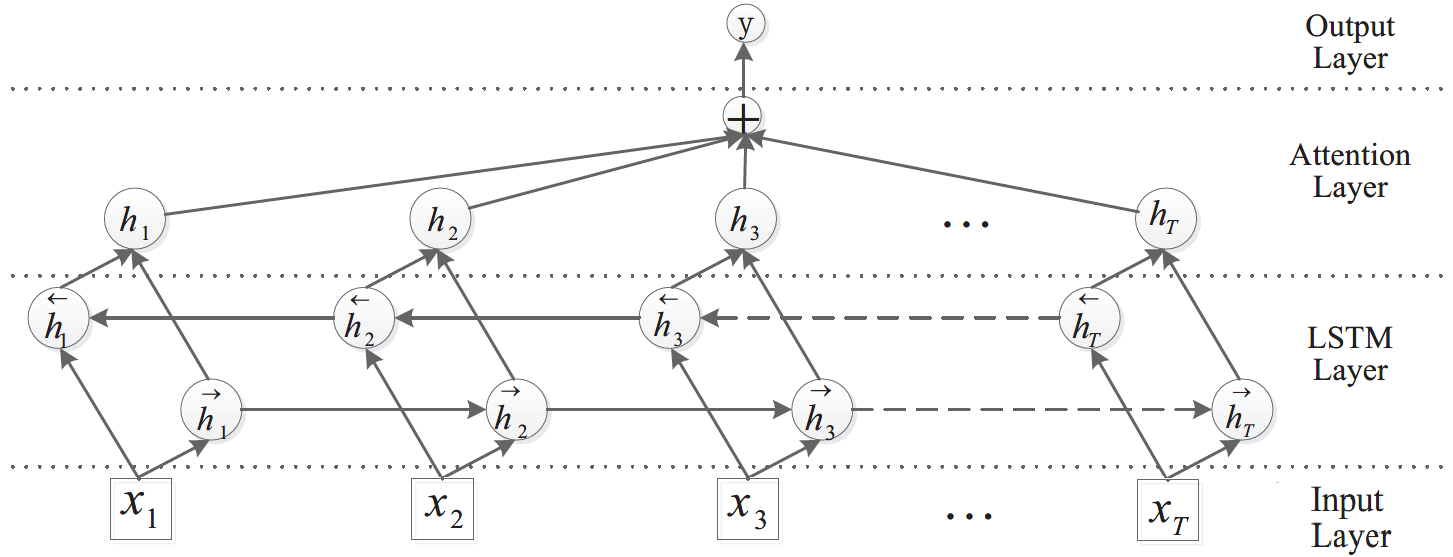}
\centering
\caption {Bi-RNNs with attention in our system}
\label{bilstm}
\end{figure}

Our end-to-end deep neural network consists of two bidirectional LSTM layers followed with attention layer. The architecture of bidirectional LSTM model with attention can be seen in Fig~\ref{bilstm}. W is the weight matrices, where $W_{xh}, W_{hh} $ are the input-hidden, hidden-hidden weight matrices, $b_{h}$ is the hidden and output bias vectors and $\sigma$ denotes the nonlinear activation function for hidden nodes. Bi-RNNs compute both forward state sequence $\stackrel{\rightarrow}{h}$ and backward state sequence
$ \stackrel{\leftarrow}{h}$, as formulated below:

\begin{equation}
  \stackrel{\rightarrow}{h_{t}}=
  \sigma(W_{x\stackrel{\rightarrow}{h}}x_{i}+W_{\stackrel{\rightarrow}{h}\stackrel{\rightarrow}{h}}\stackrel{\rightarrow}{h_{t-1}}+b_{\stackrel{\rightarrow}{h}})
\end{equation}

\begin{equation}
\stackrel{\leftarrow}{h_{t}}=
\sigma(W_{x\stackrel{\leftarrow}{h}}x_{i}+W_{\stackrel{\leftarrow}{h}\stackrel{\leftarrow}{h}}\stackrel{\leftarrow}{h_{t-1}}+b_{\stackrel{\leftarrow}{h}})
\end{equation}

\begin{equation}
  {h_{t}}= W_{\stackrel{\rightarrow}{h}}\stackrel{\rightarrow}{h_{t}}+W_{\stackrel{\leftarrow}{h}}\stackrel{\leftarrow}{h_{t}}+b
\end{equation}

Next, Let $H$ be a matrix consisting of output vectors $H=[h_{1}, h_{2}, .. h_{t} .. , h_{T}]$ that the LSTM layer produced, where T is the sentence length, then the representation $y$ of the sequence is formed by a weighted sum of these output vectors through the attention layer:

\begin{equation}
\alpha=softmax(W^T\sigma(H))
\end{equation}

\begin{equation}
y=H\alpha^T
\end{equation}

\subsection{Audio Feature Extraction}

The analysis step for most of the previous work on audio feature extraction is typically based on a specialized techniques such as Mel-frequency cepstral coefficients (MFCCs),  linear predictive coding (LPC), etc. These techniques are  widely adopted mainly due to the good linear separation of phonemes, as well as their fitness for  hidden Markov models. In our work, we tested two different input data representations (MFCCs and LPC) and observed that the MFCC coefficients tend to be superior for our case.

Every input audio sequence in our system is synchronized to the corresponding video at 30 FPS and the sampling rate for audio track is at 44100Hz. As a result, there are 1470 audio samples for each corresponding video frame. In practice, we include additional samples from the previous video frame, so that for each video frame there is enough audio data to extract two windows of 33.33ms. The input audio window is divided into 64 audio frame with 2x overlap. Therefore, we use 2.13s worth of audio as input, i.e., 1.067s of past and future samples. By choosing this value, the network is able to capture relevant effects like phoneme coarticulation without providing too much data that might lead to high inference time and overfitting. Finally, we calculate K=39 Mel-frequency cepstral coefficients to yield a total of 64$\times$39 scalars for the input audio window. In summary, the input feature vector for every video frame has 2496 dimensions.

\section{TRAINING DETAILS}
In this section, we describe the aspects relevant to training our network: how the training dataset were obtained, loss function design, and implementation details.

\subsection{Training Dataset}

\begin{figure}[h]
\centering
\includegraphics[width=0.5\textwidth]{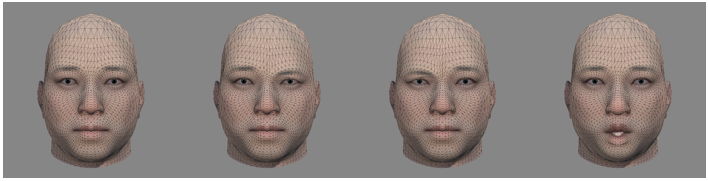}
\centering
\caption {Examples of face shape changes as certain blendshape parameter alters. From left to right: all parameters are set to zero, dimension 12 is set to 100, dimension 13 is set to 100, dimension 51 is set to 100}
\label{fig:parameter}
\end{figure}

In this work, we use 51 dimensional blendshape parameters to depict the overall shape of the whole face. Each parameter, range from 0 to 100, controls certain part of the avatar's face and the value of the parameter controls the range of actions.  Fig~\ref{fig:parameter} shows several examples of how parameters affect the face shape. We obtained the driven parameters for 3D facial model as training targets using the commercial Faceshift system, which employs depth cameras to directly capture dynamic RGB and depth information of the actor's facial movements. This system allows us to bypass the complex and expensive facial rigging while obtaining a reasonable and acceptable ground truth blendshape parameters, which animated control vertices for each frame. By processing information captured by depth camera, we can finally obtain the driving blendshape parameters to drive the pre-designed face model. These parameters - or more precisely the representations from a specific face shape - are the expecting outputs of our network when feeding a window of audio signal.

In general, our training sets consists of two parts: A female actor and a male actor. The female actor was required to read four scripts from different scenes with shorter length (each script last for about 15 minutes).  In these four datasets (denoted as F1, F2, F3 and F4), we mainly focus on the task of speech animation and just record the lower part of female actor's face. On the hand, we captured the whole facial shape changes of the male actor for intact facial animation when he was reading two longer scripts(each script last for about 30 minutes), denoted as M1 and M2. Normally, as the training data expands the inference quality will increase correspondingly. However, due to the cost of capturing high-quality training set, we choose to build a training set with acceptable size and we find that 60 minutes per actor has reach a pretty good result.

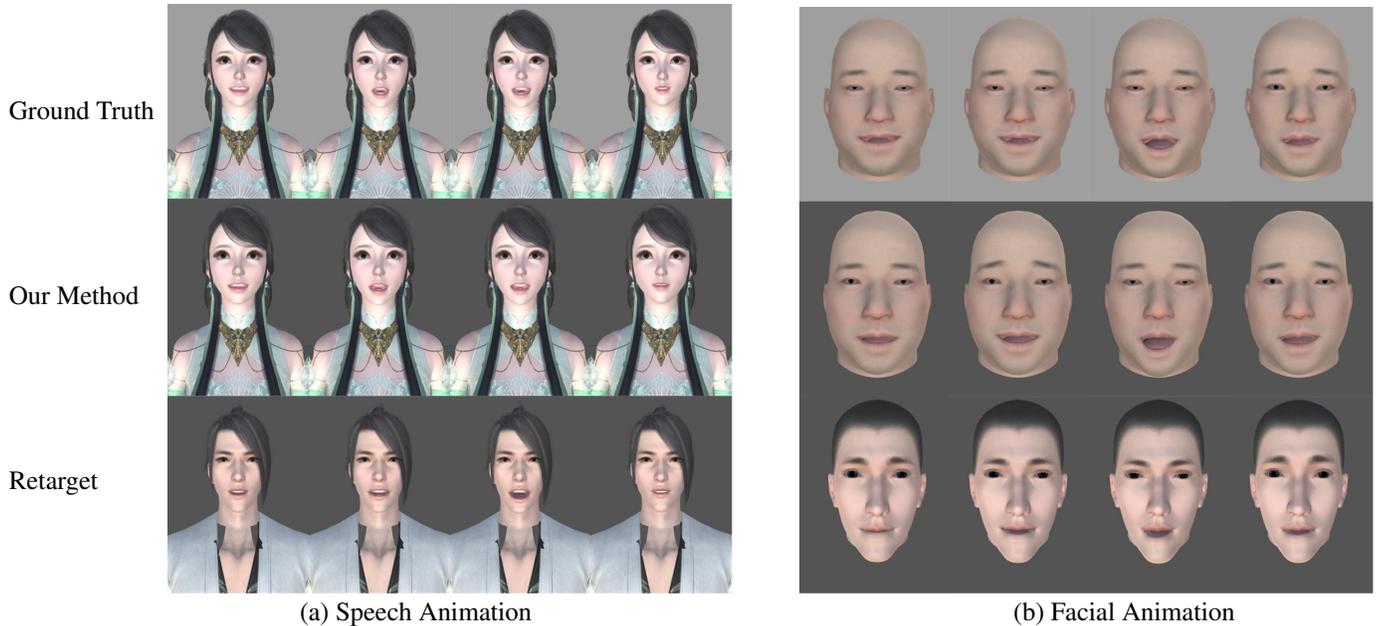
\begin{figure*}[htbp]
\centering
\subfigure{
\begin{minipage}[t]{0.5\linewidth}
\centering
\begin{picture}(500,220)
\put(60,0){\includegraphics[scale=0.28]{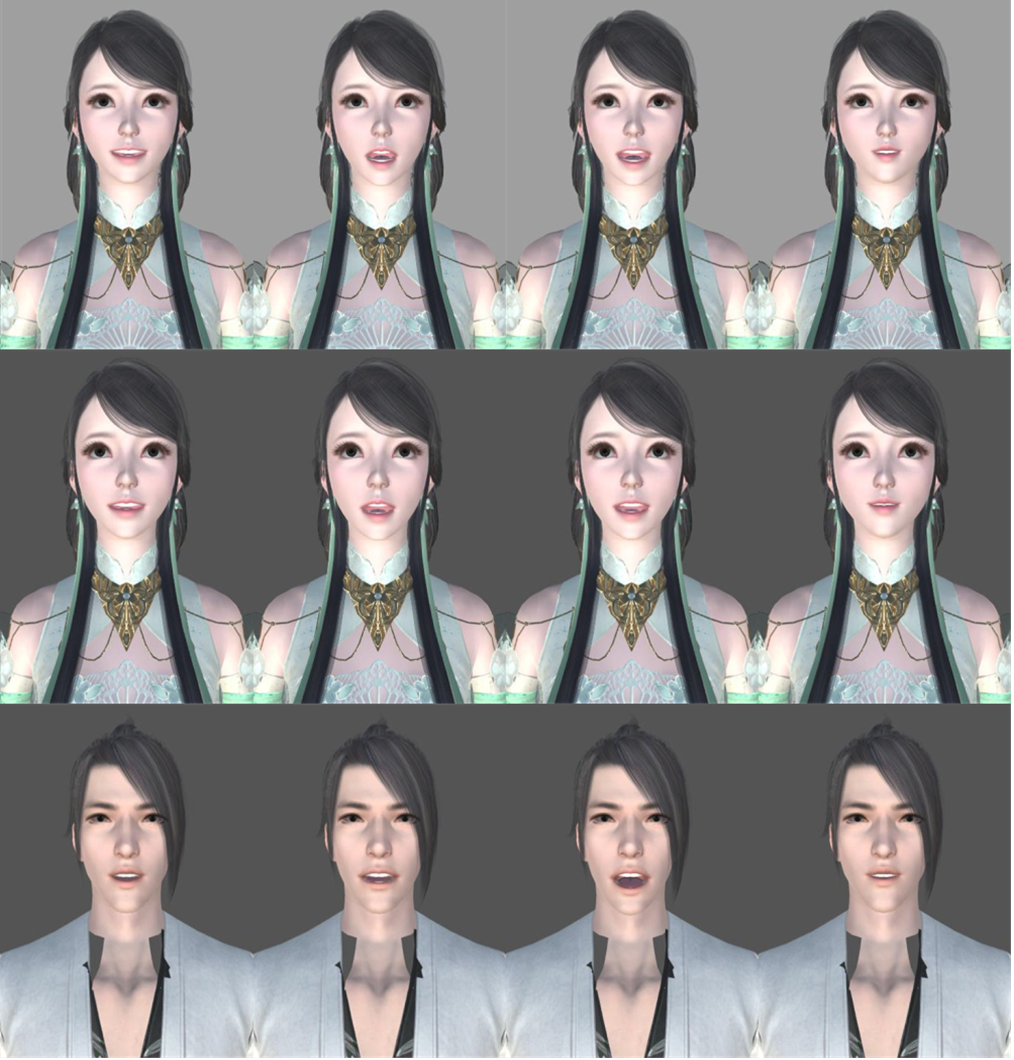}}
\put(0,180){Ground Truth}
\put(0,110){Our Method}
\put(0,40){Retarget}
\put(110,-10){(a) Speech Animation}
\put(380,-10){(b) Facial Animation}
\end{picture}
\end{minipage}%
	}%
\subfigure{
\begin{minipage}[t]{0.6\linewidth}
\centering
\includegraphics[width=3in]{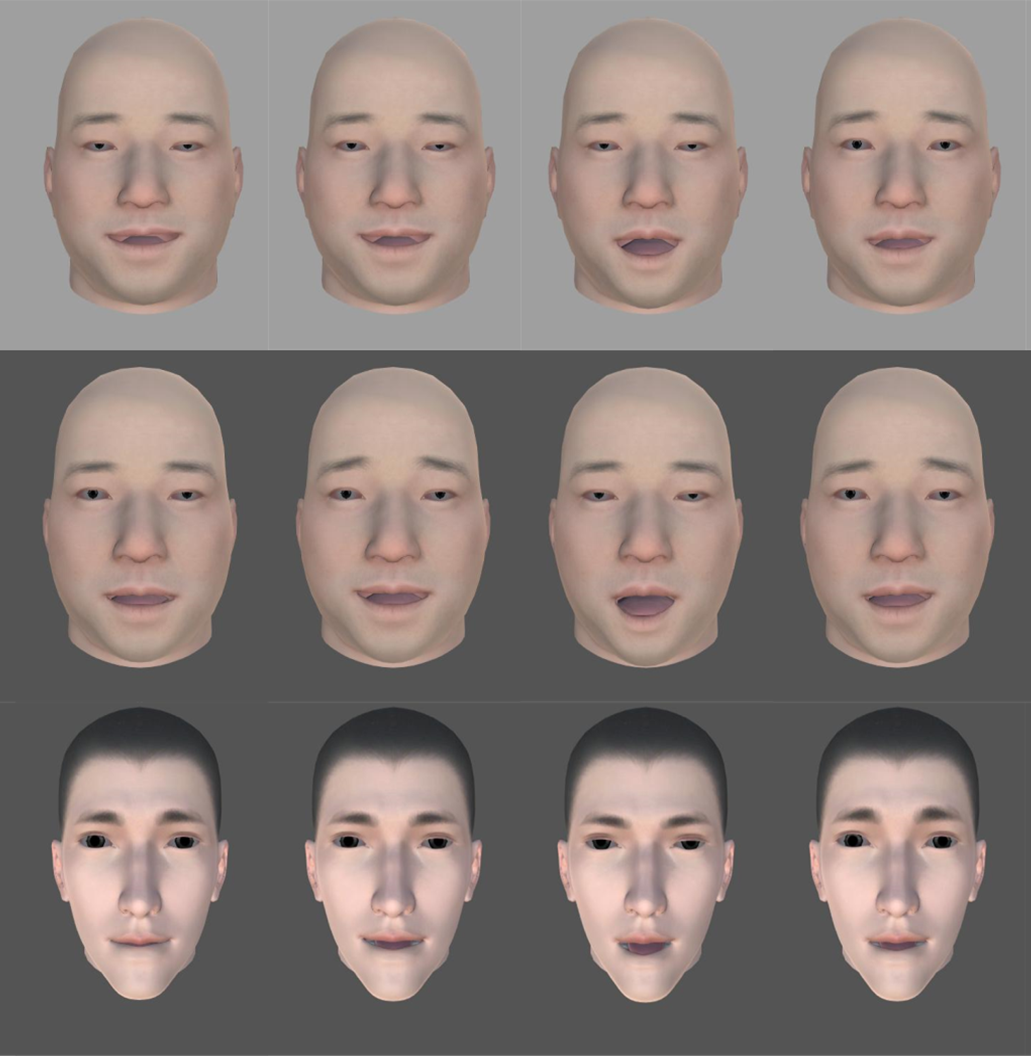}
\end{minipage}%
	}%
\centering
\caption{Comparison of held-out vision-captured ground truth of the reference speaker compared with our reference model rendered predictions. Predicted mouth regions are rendered onto the original face for speech animation (a) and predicted overall facial regions are rendered for facial animation (b).
}
\label{fig:animation}
\end{figure*}

\subsection{LOSS FUNCTION}

Motivated by the unit selection in concatenative speech synthesis and considered the ambiguity nature of the optimization target,  we design a specialized lost function that contains two losses: the target loss and the smooth loss:
\begin{equation}
  L=\frac{1}{n} \sum_{i=1}^{n}w_{1}L_{t}(y_{i},y_{i}^p) + \sum_{i=2}^{n}w_{2}L_{s}(y_{i-1}^p,y_{i}^p)
\end{equation}
where $y_{i}$ and $y_{i}^{p}$ are the desired output and the output produced by the network, while $L_{t}$ and $L_{s}$ represent the target loss and the smoothness loss,  and $w_{1}$, $w_{2}$ are target weight and smooth weight, respectively.\\
\textbf{Target loss}
We use the target loss to ensure that each output blendshape parameter is roughly correct. We  adopt Huber loss as our target loss because it is less sensitive to outliers in dataset than MSE (mean square error) loss and is able to converge faster than both MSE loss and MAE (mean absolute error) loss to the minimum:
\begin{equation}
  L_\delta(a)=\left\{
 \begin{array}{ll}
   \frac{1}{2}(y_{i}-y_{i}^p)^2                &  for|y_{i}-y_{i}^p|\leqslant \delta \\
   \delta|y_{i}-y_{i}^p|-\frac{1}{2}\delta^2   & otherwise
 \end{array} \right.
\end{equation}
$\delta$ is a hyperparameter can be tuned, which we set to 1 in practice.
\\

\textbf{Smooth Loss}
The target term is used to output a roughly correct result for animation task, but itself is  insufficient to produce facial animation with high quality:  As the coarticulation phenomenon and similarity between frames, training the network with the just target loss could lead to a considerable amount of temporal instability, as well as a weakly response to individual phonemes. Therefore, we bring in the smooth loss, taking the smoothness between adjoining selected frames into consideration. Inspired by the work of~\cite{Fan2016}, we use cosine distance to measure the smooth cost between two adjacent frame sample:
\begin{equation}
    L_{s}(y_{i}^p,y_{j}^p)=cos(y_{i}^p,y_{j}^p)=\frac{y_{i}^p\cdot y_{j}^p}{||y_{i}^p||\cdot ||y_{j}^p||}
\end{equation}
where $i ,j$ represent the $ith, jth$ animation frame.

\subsection{Implementation Details}

Our framework is implemented in Python, based on the deep learning toolkit pytorch. We train the deep models in 500 epochs, using Adam~\cite{Kingma2014} with the default parameters, where learning rate is chosen as 0.0001 and the weights are initialized with a Gaussian random distribution. Each epoch processes all training samples  in minibatches of 100. The dimension of MFCCs from audio feature extraction process is 39. We train and test our neural network on an Nvidia GTX 1080i.

\section{EVALUATION}

To assess the quality of our output, we conducted an empirical analysis evaluating our approach using both quantitative and subjective measures against several baselines. In the study, we compared the result of our method with video-based capture system from Faceshift and LSTM method~\cite{Fan2016}, as well as classical HMM-based synthesis ~\cite{Wang2015}. We choose audio clips randomly from the validation dataset of the corresponding actor as our test sets, which were not used at the training process.

\subsection{Comparison with Other Methods}
In our benchmark evaluation experiments, we evaluate all methods on our 6 self-build datasets mentioned above. We evaluate the performance of all the approaches using the same loss function, as mentioned above. HMM based method is denoted as HMM and long short-term memory networks regression is denoted as LSTM.

For the qualitative comparison, Fig~\ref{fig:animation}~(a)shows the speech animation result of one example sequence from the test set. Experiment results reveal that our output speech-driven animation quality from the input sequence is quite plausible compared to ground truth. Fig~\ref{fig:animation}~(b) is the result of our facial animation experiment. We can see that our model is able to effectively reproduce accurate face shapes on the test audio. However, there are certain limitations in our method: our network could not precisely follow blink pattern of the actor. This is a intractable problem because the speaker's blink hardly has any relationship with their speech. We are considering fixing this by adding a constant blink pattern into the framework in the future work.

For the quantitative evaluation, we tested the performance of network topologies with
different number of nodes (128, 256 and 512) and compared the performance of our method with that of HMM and LSTM, as listed in Table~\ref{t1}. The metrics in this table is root mean squared error(RMSE) of parameters over all video frames in the held-out test set:

\begin{equation}
\varepsilon=\sqrt{\frac{1}{n}\sum_{i=1}^{n}(y_{i}-y_{i}^p)^2}
\end{equation}
We can figure out from Table~\ref{t1} that our approach consistently achieves the lowest RMSE. Results show that 256 nodes per layer could get an  acceptable result for the speech animation task while 512 nodes perform superior on facial animation task. This is in accordance with our prediction because animation on entire face is more complex.

\begin{table*}[htbp]
	\centering
	\normalsize
	\caption{{RMSE when Compared with typical HMM and LSTM methods on Different Test Sets}}
	\begin{tabular}{lccccccc}
		\hline
		Algorithm & Mean &F1 & F2 & F3 & F4 & M1 & M2\\
		\hline
    HMM  & 0.3222 & 0.2345 &0.2593 &0.3211 & 0.4123 &0.3756 & 0.3301 \\
	LSTM & 0.3046 &0.2132 & 0.2410 & 0.2914 &0.4038 &0.3644 & 0.3140\\
    Our Method(128 nodes) & 0.2965 & 0.1875 &0.2365& 0.2876 &0.3714 &0.3751 & 0.3210\\
    Our Method(256 nodes) & 0.2891 & \textbf{0.1817} &0.2336& 0.2819 &\textbf{0.3648} &0.3624 & 0.3104\\
    Our Method(512 nodes) & \textbf{0.2883} & 0.1821 &\textbf{0.2328}& \textbf{0.2798} &0.3661 &\textbf{0.3589} & \textbf{0.3098}
	\end{tabular}
  \label{t1}
\end{table*}

\subsection{Retarget Test}
After our model is trained, we retarget it to a a different cartoon character. Fig. \ref{fig:animation} shows the performance of our method when generalizing the predicted animation to a new character's output face model. We can figure that basically the facial movements and expressions can match between characters.

\subsection{Run-Time Performance}
To evaluate the run-time performance of our approach (model with 256 nodes) we perform experiments on a sample of random 10 sequences from the test sets with each sequence last for about 5 seconds. Each audio window in the animation last for 33.3ms because the frame rate is 30. The experiments were performed on an Intel Xeon(R) CPU E5-2680 v4 @ 2.40GHz $\times$ 16 CPU and an Nvidia GTX 1080i GPU. In general, it requires 0.14ms on average to process one audio window and update the blendshape parameters in the 3D morphable face model on average. Besides, it need an additional 0.52ms to extract audio feature for each audio window. As is illustrated above, the other deployments in the framework do not need extra time in the inference process. Therefore, the average time for the feature extraction and a forward pass of our neural network for one audio window is 0.68ms, which makes it possible for the implementation of real-time animation.

\section{CONCLUSION}

In this paper, we introduce a deep learning approach for speech/facial animation from exclusively audio input. Our deep framework is based on bi-directional long short-term memory recurrent neural network, combining with attention mechanism. Evaluation results  demonstrate the effectiveness and generalization ability of our approach in learning the affective state and facial movement of speakers.

\section{ACKNOWLEDGMENT}
This work is supported in part by the National Key Research and Development Program of China under Grant 2017YFB1302003 and by the Fundamental Research Funds for the Central Universities(2018QNA5012).

\bibliographystyle{IEEEbib}
\bibliography{icme2019template}

\end{document}